\documentclass[11pt]{article}

\usepackage[final]{acl}

\usepackage{times}
\usepackage{latexsym}
\usepackage[T1]{fontenc}
\usepackage[utf8]{inputenc}
\usepackage{microtype}
\usepackage{inconsolata}

\usepackage{graphicx}
\usepackage{amsmath}
\usepackage{amssymb}
\usepackage{booktabs}
\usepackage{multirow}
\usepackage{array}
\usepackage{xcolor}
\usepackage{placeins}
\usepackage{stfloats}
\usepackage{float}

\title{TRIS: A Tri-Layer Retrieval Integrity Sieve Against Knowledge Poisoning}

\author{
 \textbf{Muhaimin Bin Munir\textsuperscript{1}},
 \textbf{Akib Jawad Ononto\textsuperscript{1}},
 \textbf{ Nazia Shehnaz Joynab \textsuperscript{1}},
\\
 \textbf{ Bhavani Thuraisingham\textsuperscript{1}},
 \textbf{Latifur Khan\textsuperscript{1}},
\\
\\
 \textsuperscript{1}University of Texas at Dallas,
\\
 \small{
   \textbf{Correspondence:} \href{mailto:muhaimin.binmunir@utdallas.edu}{muhaimin.binmunir@utdallas.edu}
 }
}

\begin{document}
\maketitle

\begin{abstract}
Retrieval-Augmented Generation (RAG) grounds large language models in external corpora, but implicit trust in retrieved documents creates a critical attack surface: PoisonedRAG shows that a handful of crafted passages can dominate dense retrieval and steer generation toward attacker-chosen answers. We present the Tri-Layer Sieve, a middleware defense that sanitizes retrieved evidence through cross-embedding-space clustering with an independent judge model, structural filtering of trigger--payload artifacts, and LLM consistency verification. The design exploits a key weakness of retrieval-stage poisoning: a single document must satisfy one embedding geometry, one internal Trigger--Payload structure, and one generation objective --- rarely all three simultaneously, a fragility that persists even against an adaptive attacker who paraphrases around it. On Natural Questions, HotpotQA, and MS-MARCO with Contriever retrieval ($k=50$), the Sieve reduces black-box Attack Success Rate from 67.0/87.0/64.0\% to 3.0/14.0/4.0\%, mitigates white-box HotFlip attacks from $\sim$74\% to 27.8\% on NQ with Layer~3 enabled, and drives poisoned-document MRR to $0.000$, while restoring clean accuracy from $13$--$33\%$ under attack to $58$--$76\%$. Under an architecture-aware adversary who paraphrases triggers to evade the structural filter, enabling the consistency layer halves adaptive ASR (32.0\%~$\to$~15.0\% on NQ) while raising clean accuracy by 18 points, at an added latency of $\sim\!16$--$19$\,s/query under live retrieval.
\end{abstract}

%-------------------------------------------------------------------------------
\section{Introduction}
%-------------------------------------------------------------------------------

Large language models are limited by their parametric memory: they cannot be updated continuously, hallucinate plausible-but-false statements, and struggle with domain-specific facts. Retrieval-Augmented Generation (RAG) \citep{lewis2020retrieval} addresses this by grounding outputs in external corpora retrieved at inference time, and is now the de facto architecture for knowledge-intensive search, open-domain question answering, and enterprise deployments \citep{gao2023retrieval}.

This advantage introduces a corresponding security flaw: a RAG system consumes mutable external data at inference time and treats every retrieved passage as trustworthy by default. \citet{zou2025poisonedrag} (PoisonedRAG) showed this implicit trust can be exploited --- inserting as few as five crafted documents into a corpus such as MS~MARCO \citep{nguyen2016msmarco} or a Wikipedia snapshot suffices to dominate dense retrieval and force attacker-chosen misinformation, even when the model knows the correct answer parametrically. Follow-up work extends the threat model to jamming \citep{shafran2024jamming}, agent memory poisoning \citep{chen2024agentpoison}, backdoor triggers \citep{cheng2024trojanrag}, and benchmarks \citep{ma2025benchmarking}, establishing retrieval-stage poisoning as a distinct, under-defended threat surface.

Existing defenses are insufficient for two reasons. \emph{First}, grounding-verification and hallucination-detection methods \citep{huang2023hallucination,song2024veriscore} act on the \emph{output} of generation rather than the integrity of retrieved evidence, and so cannot prevent the \emph{retrieval dominance} PoisonedRAG exploits. \emph{Second}, defenses targeting poisoned retrieval --- trust-weighted scoring \citep{zhou2025trustrag}, certified isolate-and-aggregate decoding \citep{xiang2024robustrag}, and query paraphrasing \citep{zou2025poisonedrag} --- each address only one constraint a poisoned document must satisfy, and can trade sharply against utility on complex queries when tuned aggressively (Section~\ref{sec:fair-baselines}).

We argue that retrieval-stage poisoning succeeds precisely because the attacker must simultaneously satisfy several distinct objectives --- dense-retrieval similarity under a specific encoder, internal coexistence of a retrieval-optimized \emph{Trigger} and a generation-optimized \emph{Payload}, and factual plausibility against the LLM's parametric knowledge --- and that defenses should exploit this multi-objective fragility rather than detect surface artifacts.

\textbf{Contributions.} We propose \emph{TRIS} (the Tri-Layer Retrieval Integrity Sieve)\footnote{Code, evaluation scripts, and poison-generation data: \url{https://github.com/akibjawad14/tris}}, a middleware defense that intercepts retrieved documents between the retriever and the generator and applies three orthogonal filters --- cross-embedding-space clustering, structural trigger--payload detection, and LLM consistency verification --- each targeting one of the three constraints a successful retrieval-stage poison must simultaneously satisfy. Against the PoisonedRAG suite on Natural Questions \citep{kwiatkowski2019natural}, HotpotQA \citep{yang2018hotpotqa}, and MS-MARCO \citep{nguyen2016msmarco} with Contriever \citep{izacard2021contriever}, TRIS reduces black-box ASR by an order of magnitude on NQ and MS-MARCO while recovering $41$--$45$ points of clean accuracy over the attacked baseline, and outperforms TrustRAG \citep{zhou2025trustrag} on MS-MARCO ($4\%$ vs.\ $20\%$ ASR); against TrustRAG and RobustRAG \citep{xiang2024robustrag} at fair operating points it is competitive rather than dominant, at substantially lower verification cost (Section~\ref{sec:fair-baselines}). We further contribute layer-wise ablations, an ablation isolating Layer~1's embedding geometry, and an empirical evaluation of architecture-aware (Level 1) adaptive adversaries under live retrieval and a forced-top worst case. We use TRIS and the Tri-Layer Sieve interchangeably.

%-------------------------------------------------------------------------------
\section{Background and Related Work}
%-------------------------------------------------------------------------------
\subsection{Retrieval-Augmented Generation}
\label{sec:rag-bg}

RAG \citep{lewis2020retrieval} augments LLMs with non-parametric memory: a query $q$ is embedded by $f_Q$ and the top-$k$ documents in a corpus $\mathcal{D}$ are retrieved by $s(q,d) = \mathrm{sim}(f_Q(q), f_D(d))$, after which an LLM conditions on that evidence. Dense retrievers such as DPR \citep{karpukhin2020dense} and Contriever \citep{izacard2021contriever} are now standard. RAG offers stronger factuality than parametric-only LLMs \citep{gao2023retrieval}, but trusts retrieved evidence by default, opening a new attack surface.

\subsection{Knowledge Poisoning Attacks}

Classical data poisoning targets training pipelines \citep{geiping2021witchesbrew,steinhardt2017certified,gu2017badnets}. Retrieval-stage poisoning is a newer threat that requires only corpus access \citep{zou2025poisonedrag,shafran2024jamming,edemacu2025ragdefender}.

\paragraph{PoisonedRAG.} \citet{zou2025poisonedrag} give the first systematic demonstration of retrieval-stage poisoning for LLM-based RAG. Each poisoned document is $d_{\mathrm{adv}} = S \oplus I$: a retrieval-optimized Trigger $S$ and a generation-optimized Payload $I$. Black-box $S$ repeats the query; white-box $S$ is gradient-optimized via HotFlip \citep{ebrahimi2018hotflip} or universal triggers \citep{wallace2019universal}. Five poisons per query suffice for ASR$>$$90\%$.

\subsection{Existing Defenses}
\label{sec:related-defenses}

\textbf{Output-side methods} \citep{song2024veriscore} detect post-generation mismatches but cannot prevent retrieval dominance. \textbf{TrustRAG} \citep{zhou2025trustrag} re-ranks via a single learned trust score (with an optional LLM-consistency check); effective on simple attacks, weaker on MS-MARCO (Section~\ref{sec:results}). TRIS differs architecturally, not just numerically: three independent off-the-shelf checks rather than one learned scorer, each targeting a different constraint, evaluated against an explicit adaptivity taxonomy (Section~\ref{sec:adaptive-threat}). We scope that difference precisely: Layer~2 targets verbatim and near-duplicate injection --- the common case --- where TrustRAG has no dedicated structural filter (ROUGE-L overlap is its closest analogue). Under paraphrased triggers Layer~2 fires on roughly zero documents per query and TrustRAG's more lenient check catches more, so we claim Layer~2 for the common attack, not as a paraphrase defense (Section~\ref{sec:adaptive-eval}).

\textbf{RobustRAG} \citep{xiang2024robustrag} gives isolate-and-aggregate decoding certified against $k$-corruption, but strict isolation harms clean accuracy on multi-hop queries. \textbf{Query paraphrasing} helps only marginally, since dense retrievers map paraphrases into similar embedding regions \citep{zou2025poisonedrag}. \textbf{Perplexity detection} fails because LLM-generated payloads are often \emph{more} fluent than genuine web text \citep{shafran2024jamming}. \textbf{Prompt injection} \citep{greshake2023notwhat} is related but distinct: our payloads encode a false fact, not an instruction.

%-------------------------------------------------------------------------------
\section{Threat Model}
\label{sec:threat-model}
%-------------------------------------------------------------------------------

We consider an adversary whose goal is to manipulate the output of a RAG system by injecting adversarial content into its external corpus. Our threat model follows \citet{zou2025poisonedrag} and generalizes to both structured and unstructured corpora common in real deployments. Figure~\ref{fig:poisonedrag-threat-model} in Appendix~\ref{sec:appendix-fig1} traces this attack pipeline end to end for a single example query.

\subsection{System and Adversary Model}

A RAG pipeline consists of (i) a retriever $\mathcal{M}_{\mathrm{ret}}$ that embeds $q$ and retrieves the top-$k$ from corpus $\mathcal{D}$, and (ii) a generator $\mathcal{M}_{\mathrm{gen}}$ that conditions its output on $\mathcal{D}_k$. The generator trusts retrieved evidence by default, matching production deployments over mutable corpora.

The adversary may (a) \textbf{inject documents} into $\mathcal{D}$, (b) \textbf{craft adversarial triggers} that maximize similarity with $q^*$ under $\mathcal{M}_{\mathrm{ret}}$ (verbatim query repetition in the black-box setting; HotFlip \citep{ebrahimi2018hotflip} gradient optimization in the white-box setting), and (c) \textbf{embed malicious payloads} that express $y_{\mathrm{adv}}$ as an authoritative encyclopedic claim. These capabilities instantiate the Split-and-Merge construction of \citet{zou2025poisonedrag}: a poisoned document concatenates a retrieval-optimized Trigger with a generation-optimized Payload, independently optimizing retrieval dominance and output steering.

The adversary's goals are to dominate retrieval for $q^*$, override the LLM's internal knowledge, and induce a targeted output $y_{\mathrm{adv}} \neq y_{\mathrm{true}}$ \citep{zou2025poisonedrag,shafran2024jamming}. The defender (i) cannot retrain $\mathcal{M}_{\mathrm{ret}}$ or $\mathcal{M}_{\mathrm{gen}}$, (ii) cannot modify $\mathcal{D}$ beyond lightweight preprocessing, and (iii) must operate as a middleware layer --- constraints reflecting deployed enterprise settings.

\subsection{Adaptive Adversaries}
\label{sec:adaptive-threat}

A reviewer might reasonably ask whether TRIS (Section \ref{sec:defense}) withstands attackers who know about it. We distinguish four levels of adaptivity, of which the evaluated PoisonedRAG attacks are Level 0:

\textbf{L0 (static)}: the attacker is unaware of the defense --- PoisonedRAG black-box and white-box, our headline setting. \textbf{L1 (architecture-aware)}: the attacker knows TRIS is deployed but not its parameters, and may paraphrase rather than repeat $q$ (defeating Layer~2) and write encyclopedic-style payloads (degrading Layer~3); Layer~1's geometric check still applies. \textbf{L2 (judge-aware)}: the attacker also knows $\mathcal{J}$ and jointly optimizes for rank under $\mathcal{R}$ \emph{and} majority-cluster membership under $\mathcal{J}$ --- a conjunction constrained by the geometric independence of the two spaces, and compounded by ensembling or rotating $\mathcal{J}$. \textbf{L3 (full white-box)}: the attacker also knows Layer~2's thresholds and $y_{\mathrm{int}}$, and must craft a payload contradicting $y_{\mathrm{int}}$ that the LLM finds equally plausible.

We now evaluate Level 1 empirically (Section~\ref{sec:adaptive-eval}); Levels 2 and 3 remain conceptual only, and closing that gap is the most important remaining limitation of our evaluation (Section~\ref{sec:limits}). A core claim of this work --- now supported empirically for Level 1 --- is that defeating an $n$-layer orthogonal defense requires the attacker to satisfy the conjunction of $n$ constraints in non-aligned objective landscapes, which substantially enlarges the attack search space even without certified bounds.

% We do not yet evaluate Levels 2 or 3; this is the most important gap in our evaluation (Section~\ref{sec:limits}). A core conceptual claim of this work is that defeating an $n$-layer orthogonal defense requires the attacker to satisfy the conjunction of $n$ constraints in non-aligned objective landscapes, which substantially enlarges the attack search space even without certified bounds.

\subsection{Out of Scope}
We exclude (i) training-set poisoning, (ii) prompt-injection attacks where the user is malicious, and (iii) attacks that require control of LLM weights.

%-------------------------------------------------------------------------------
\section{The Tri-Layer Sieve}
\label{sec:defense}
%-------------------------------------------------------------------------------

The Tri-Layer Sieve is a defense-in-depth middleware between the retriever and the generator. Its central principle is that no retrieved document is trusted at face value: every passage is checked against up to three orthogonal filters before reaching the generator, with the third invoked adaptively rather than on every document. The three layers target the three constraints that retrieval-stage poisoning must simultaneously satisfy: retrieval-side embedding similarity, internal trigger--payload coexistence, and factual alignment with parametric knowledge.

\begin{figure*}[t]
    \centering
    \includegraphics[clip, trim=0cm 9cm 0cm 0cm, width=0.85\textwidth]{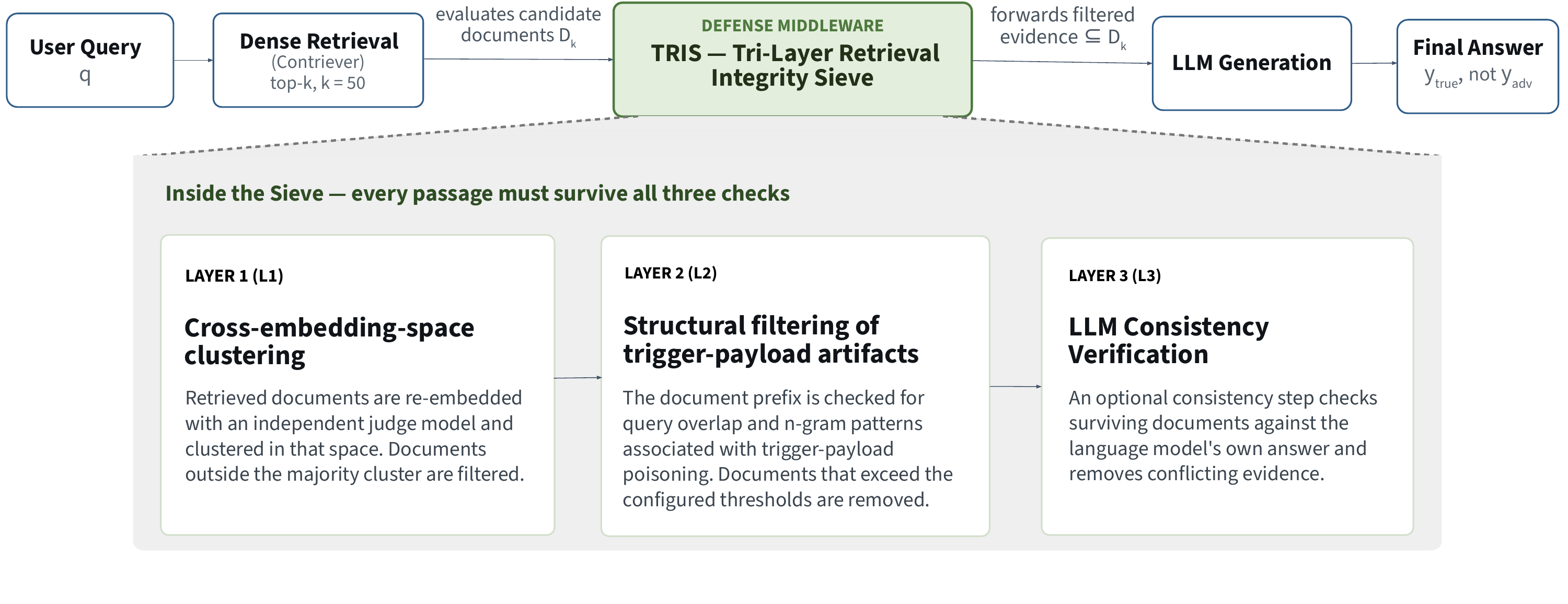}
    \caption{\textbf{High-level architecture of the Tri-Layer Sieve.} The Sieve sits between the retriever and the generator, intercepting candidate documents and forwarding only those that pass the applicable filters (Layer~3 fires adaptively, not on every document). The defense requires no modification of the retriever or LLM.}
    \label{fig:tri-layer-sieve-architecture}
\end{figure*}

Figure~\ref{fig:tri-layer-sieve-architecture} diagrams the resulting pipeline.

\subsection{Layer 1: Semantic Clustering via Model Diversity}

Poisoned documents are optimized against a specific retriever encoder, and adversarial trigger optimization \citep{ebrahimi2018hotflip,wallace2019universal} exploits the geometry of that encoder rather than true semantic content. Layer~1 is therefore designed as \emph{majority-keep in an architecturally independent geometry}: rather than assuming poisons fail to cluster near the query under a second encoder, it assumes nothing about where they land and simply retains the majority cluster. We frame this as a design property --- it drops the requirement that poisons form a tight, separable cluster in the retriever's own space --- not as a source of empirical gain; Appendix~\ref{sec:appendix-judge} tests the geometry directly and finds it is not the differentiating factor.

Layer~1 re-embeds the top-$k$ retrieved documents with a judge model $\mathcal{J}$ that is architecturally distinct from the retriever $\mathcal{R}$ --- we use Sentence-BERT \texttt{all-MiniLM-L6-v2} \citep{reimers2019sentencebert}. After $L_2$-normalization we run K-Means with a small cluster count $C$ and identify the majority cluster; we sweep $C \in \{2,3,5\}$ in Section~\ref{sec:results} and find $C=3$ the default sweet spot. Poisons typically appear as isolated micro-clusters, high-density off-distribution clusters, or low-similarity outliers. Documents outside the majority cluster are dropped, with a fallback retaining all documents when no clear majority exists.

\subsection{Layer 2: Content-Based Structural Filtering}

Layer~2 targets the structural footprint of Split-and-Merge: Trigger and Payload need not share semantic content, but must coexist in one document, producing lexical irregularities at the prefix. For each surviving document we compute token-level Jaccard similarity between $q$ and the first $T=20$ tokens, bigram/trigram overlap between $q$ and the prefix, and length/repetition patterns; if Jaccard or $n$-gram overlap exceeds a configurable threshold (we use $0.8$), the document is discarded. Unlike perplexity filtering --- which fails because LLM payloads are highly fluent \citep{shafran2024jamming} --- this detects the poison's \emph{functional decomposition}. Evasion is cheap at the retrieval stage, however: Section~\ref{sec:adaptive-eval} shows a paraphrased trigger keeps near-identical rank while falling below the threshold, shifting the burden onto Layers 1 and 3. Layer~2 costs $\sim\!8$\,ms per query.

% this layer detects the \emph{functional decomposition} of the poison, which is hard to evade without sacrificing retrieval dominance. 

\subsection{Layer 3: LLM Consistency Checking}

Layer~3 verifies semantically by prompting an LLM (the generator $\mathcal{M}_{\mathrm{gen}}$ in self-knowledge mode, or a separate verifier) to judge each document in isolation, which defeats the dominance effect when many retrieved passages repeat the same poisoned claim; LLMs retain strong parametric priors even when retrieval corrupts the final answer \citep{longpre-etal-2021-entity}. The verifier (i) answers $q$ without context to capture its parametric belief $y_{\mathrm{int}}$, (ii) summarizes each document's claim relevant to $q$, and (iii) judges it compatible with $y_{\mathrm{int}}$, contradictory, or uncertain; only strong contradictions without evidence of a legitimate update are dropped. Crucially, if the verifier is not confident --- or the call fails or returns a malformed response --- the document \textbf{fails open} and is retained, so Layer~3 can never remove a document it cannot judge, and verifier unavailability degrades it to a no-op rather than discarding evidence under uncertainty (this underlies the valid-response-only rows in Section~\ref{sec:setup} and Table~\ref{tab:ablation}). Two invocation modes exist: \emph{always-on} (used for L3 ablation rows) and \emph{adaptive}, invoked only when Layers~1 and 2 disagree.

\subsection{Composition, Default Configuration, and Engineering}

The three layers are \emph{complementary in coverage} rather than strictly additive: triggers that evade Layer~1 are typically caught by Layer~2's prefix overlap check, and payloads that survive Layer~2 trigger contradictions in Layer~3. Composing all three does not strictly dominate every pairwise combination, however, since Layer~3 occasionally discards clean passages whose phrasing differs from the LLM's prior.

We therefore recommend \textbf{a single default: L1+L2 with Layer~3 in adaptive mode}, invoked only where Layers~1 and 2 disagree. It attains the lowest black-box ASR ($3.0\%$ on NQ) at $\sim\!12$\,ms/query, and adaptive invocation recovers most of Layer~3's white-box and architecture-aware robustness without paying always-on latency on every query. Deployments with a known threat profile should override this: white-box-dominant settings use \textbf{L1+L3 always-on} ($27.8\%$ white-box ASR),and settings expecting trigger-paraphrasing adversaries should enable Layer~3 always-on instead (Section~\ref{sec:adaptive-eval} gives the resulting ASR, accuracy, and latency). Abstract numbers correspond to the L1+L2 default unless stated. TRIS is a drop-in Python module wrapping the retrieval call: model-agnostic, batched-cache friendly, falling back to unfiltered $\mathcal{D}_k$ with a logged warning when all documents are filtered.

%-------------------------------------------------------------------------------
\section{Experimental Setup}
\label{sec:setup}
%-------------------------------------------------------------------------------

\paragraph{Datasets and attack.} We evaluate on Natural Questions \citep{kwiatkowski2019natural}, HotpotQA \citep{yang2018hotpotqa}, and MS-MARCO \citep{nguyen2016msmarco}. For each we sample $100$ target queries; the Poison Factory generates five poisoned passages per query via Split-and-Merge \citep{zou2025poisonedrag} with a counterfactual $y_{\mathrm{adv}}$, a poisoning ratio well under $1\%$ of the corpus. We consider \emph{black-box LM-targeted} (trigger = query repeated with light paraphrasing) and \emph{white-box HotFlip} (gradient-optimized against Contriever), with $k=50$ and $10$ iterations of $M=10$ queries.

\paragraph{Adaptive-adversary experiments.} For Section~\ref{sec:adaptive-eval} we construct three additional Level-1 attack variants: \emph{trigger-paraphrased} (trigger paraphrased rather than repeated verbatim), \emph{payload-diversified} (five independently-phrased payloads rather than a shared template), and \emph{full adaptive} (both combined). The live-retrieval comparison (Table~\ref{tab:adaptive-live}) follows the same protocol as above ($n=100$, $k=50$) on NQ and HotpotQA. The forced-top worst case (Table~\ref{tab:forced-top}) pins poisons to the top of context to isolate defense behavior from retrieval competition ($n=30$, NQ). The extended injection sweep (Table~\ref{tab:inject-ext}) uses a reduced sample ($n=25$, NQ) due to the added query volume at each injection depth.

\paragraph{Fair-baseline experiments.} The RobustRAG threshold sweep (Table~\ref{tab:robustrag-alpha}) re-runs the black-box attack at $n=30$ on both datasets for each $\alpha \in \{0.1, 0.2, 0.5\}$ plus majority aggregation; the TrustRAG comparison with its LLM check enabled (Table~\ref{tab:trustrag-adaptive}) uses $n=30$ on NQ under the two adaptive variants above. The Layer-1 judge-geometry ablation (Appendix~\ref{sec:appendix-judge}) runs at full scale ($n=100$, $k=50$, both datasets), swapping only Layer~1's embedding model. Reported $p$-values are two-sided two-proportion $z$-tests over the $n=100$ per-condition query outcomes, treating queries as independent.

\paragraph{Models.} Contriever \citep{izacard2021contriever} retriever (HuggingFace, FP16); Sentence-BERT \texttt{all-MiniLM-L6-v2} \citep{reimers2019sentencebert} judge; FAISS \texttt{IndexFlatIP} \citep{johnson2019faiss} on $L_2$-normalized embeddings; GPT-3.5-turbo \citep{openai2023finetuning} generator (temperature $0$). We cross-check with Llama-2-7B-chat \citep{touvron2023llama2} and obtain consistent trends. NVIDIA A10G GPUs. Commercial API calls occasionally failed transiently (timeouts, rate limits); Layer~3 fails open on such failures (Section~\ref{sec:defense}). Where failures affected part of a condition we compute the affected numbers over valid responses only, marking them $\sim$ (Table~\ref{tab:ablation}); where an entire cell was lost we omit the cell (Table~\ref{tab:heatmap}).

\paragraph{Baselines.} (i) No Attack (Clean); (ii) Vanilla RAG (Attacked); (iii) TrustRAG \citep{zhou2025trustrag}, run in its base configuration --- trust-score re-ranking with the optional LLM-consistency check \emph{disabled} --- for Table~\ref{tab:main}, and with that check \emph{enabled} for the adaptive comparison in Table~\ref{tab:trustrag-adaptive}; (iv) RobustRAG \citep{xiang2024robustrag} with keyword-based isolate-and-aggregate decoding, using majority aggregation for Table~\ref{tab:main} and swept over the aggregation threshold $\alpha \in \{0.1, 0.2, 0.5\}$ in Section~\ref{sec:fair-baselines}. Our RobustRAG implementation extracts keywords with a regex-plus-stopword heuristic over the retrieved documents, a faithful but simplified stand-in for the authors' LLM-based extraction step; we flag this as a possible source of divergence from their reported numbers.

\paragraph{Metrics.} \textbf{ASR} ($\downarrow$): attack succeeds if the generated answer contains $y_{\mathrm{adv}}$ (case-normalized, article-stripped exact match). \textbf{CleanAcc} ($\uparrow$): the answer contains the gold string under the same normalization. \textbf{Retrieval metrics} on NQ: Recall@5, Recall@50, Clean MRR (over benign passages), Poisoned MRR (over adversarial passages). PoisonedMRR$=1.000$ means every poison ranks first; $0.000$ means full removal.

\subsection{Evaluation Protocol}
\label{sec:protocol}

Unless a table caption states otherwise, each cell aggregates $10$ iterations of $M=10$ queries ($100$ trials). The reduced-scale tables --- Tables~\ref{tab:robustrag-alpha}, \ref{tab:trustrag-adaptive}, \ref{tab:forced-top}, and~\ref{tab:inject-ext} --- print their own $n$ in the caption and are single runs at that scale, not $10$-iteration aggregates; differences of a few points in those tables should not be read as resolved. Within each iteration we re-sample queries, re-generate poisons, and re-run the pipeline, capturing variance from poison generation, retrieval order, and LLM decoding (despite temperature $0$, GPT-3.5-turbo shows residual output variation). Pairwise differences $>\!3$ points are stable across re-runs; smaller ones are within noise. Clean Recall@5 of $0.180$ reflects our $100$-query Wikipedia subset.

%-------------------------------------------------------------------------------
\section{Results}
\label{sec:results}
%-------------------------------------------------------------------------------

\subsection{End-to-End Defensive Efficacy}

Table~\ref{tab:main} reports clean accuracy and ASR across all three datasets under the black-box LM-targeted attack. Vanilla RAG is catastrophically vulnerable: ASR ranges from $64.0\%$ (MS-MARCO) to $87.0\%$ (HotpotQA), and clean accuracy collapses from $74$--$82\%$ to $13$--$33\%$ under attack. TRIS restores robustness across all three datasets: ASR drops to $3.0\%$ on NQ, $14.0\%$ on HotpotQA, and $4.0\%$ on MS-MARCO, while CleanAcc recovers by $41$, $45$, and $44$ points respectively over the attacked system and matches or exceeds every baseline on all three datasets. The residual gap to the no-attack ceiling is $2$ points on NQ and $6$ on MS-MARCO, but $16$ on HotpotQA, where Layer~3 over-filters multi-hop evidence.

\paragraph{Comparison with prior defenses.} The picture is dataset-dependent and worth reporting honestly. RobustRAG \citep{xiang2024robustrag} attains the lowest HotpotQA ASR ($1.0\%$) here, but at $5.0\%$ clean accuracy --- the aggressive end of its trade-off rather than a property of the method; Section~\ref{sec:fair-baselines} sweeps its threshold and we rest no claim on this point. TrustRAG \citep{zhou2025trustrag} is the strongest competitor, matching TRIS on NQ ($2\%$ vs.\ $3\%$ ASR at identical CleanAcc) and slightly beating it on HotpotQA ($11\%$ vs.\ $14\%$ ASR at $57\%$ vs.\ $58\%$ CleanAcc), with its LLM-consistency check disabled here (Table~\ref{tab:trustrag-adaptive} enables it); on MS-MARCO it lags substantially ($20\%$ vs.\ $4\%$ ASR, $16$ points lower CleanAcc). Comparable on NQ, marginally behind on HotpotQA at within-noise margins (Section~\ref{sec:protocol}), substantially ahead on MS-MARCO: we frame the contribution as the strongest \emph{worst-case} profile across benchmarks, not the best result on each.

\begin{table*}[t]
\centering
\small
\begin{tabular}{lcccccc}
\toprule
& \multicolumn{2}{c}{\textbf{NQ}} & \multicolumn{2}{c}{\textbf{HotpotQA}} & \multicolumn{2}{c}{\textbf{MS-MARCO}} \\
\cmidrule(lr){2-3}\cmidrule(lr){4-5}\cmidrule(lr){6-7}
\textbf{Defense} & CleanAcc$\uparrow$ & ASR$\downarrow$ & CleanAcc$\uparrow$ & ASR$\downarrow$ & CleanAcc$\uparrow$ & ASR$\downarrow$ \\
\midrule
No Attack (Clean) & 76.0\% & --- & 74.0\% & --- & 82.0\% & --- \\
Vanilla RAG (Attacked) & 33.0\% & 67.0\% & 13.0\% & 87.0\% & 32.0\% & 64.0\% \\
RobustRAG~\citep{xiang2024robustrag} & 63.0\% & 2.0\% & 5.0\% & 1.0\% & 76.0\% & 5.0\% \\
TrustRAG~\citep{zhou2025trustrag} & 74.0\% & 2.0\% & 57.0\% & 11.0\% & 60.0\% & 20.0\% \\
Tri-Layer Sieve (Ours) & \textbf{74.0\%} & \textbf{3.0\%} & \textbf{58.0\%} & \textbf{14.0\%} & \textbf{76.0\%} & \textbf{4.0\%} \\
\bottomrule
\end{tabular}
\caption{\textbf{End-to-end defensive efficacy} on NQ, HotpotQA, and MS-MARCO under PoisonedRAG black-box LM-targeted attack ($n=100$ queries per dataset, $k=50$, Contriever, GPT-3.5-turbo). The Sieve substantially outperforms TrustRAG on MS-MARCO ($4\%$ vs.\ $20\%$ ASR) while matching CleanAcc. Baselines are shown here in the single configuration used at submission time --- TrustRAG with its LLM-consistency check disabled, RobustRAG with majority aggregation; Section~\ref{sec:fair-baselines} re-examines both across their tuning knobs, and those results, not this table alone, are the basis for our comparative claims.}
\label{tab:main}
\end{table*}

\subsection{Fair Baseline Operating Points}
\label{sec:fair-baselines}

Table~\ref{tab:main} fixes each baseline at one configuration, but both expose a tuning knob that moves them along an ASR--utility frontier, so a single-point comparison can flatter either system. Table~\ref{tab:robustrag-alpha} sweeps RobustRAG's threshold $\alpha$, and Table~\ref{tab:trustrag-adaptive} re-evaluates TrustRAG with its LLM-consistency check enabled. Both are reduced-scale ($n=30$): enough to locate a frontier, not to resolve a few points.

\begin{table}[t]
\centering
\small
\setlength{\tabcolsep}{4pt}
\begin{tabular}{lcccc}
\toprule
& \multicolumn{2}{c}{\textbf{NQ}} & \multicolumn{2}{c}{\textbf{HotpotQA}} \\
\cmidrule(lr){2-3}\cmidrule(lr){4-5}
\textbf{Setting} & ASR$\downarrow$ & Clean$\uparrow$ & ASR$\downarrow$ & Clean$\uparrow$ \\
\midrule
Majority$^\dagger$ & 16.7\% & 23.3\% & 13.3\% & 53.3\% \\
$\alpha = 0.1$ & 10.0\% & 36.7\% & 6.7\% & 63.3\% \\
$\alpha = 0.2$ & \textbf{6.7\%} & \textbf{36.7\%} & \textbf{3.3\%} & \textbf{66.7\%} \\
$\alpha = 0.5$ & 6.7\% & 23.3\% & 3.3\% & 60.0\% \\
\bottomrule
\end{tabular}
\caption{\textbf{RobustRAG operating-point frontier} (black-box LM-targeted attack, $n=30$, reduced scale). Sweeping the keyword-aggregation threshold $\alpha$ moves RobustRAG along an ASR--utility frontier; $\alpha=0.2$ dominates the majority-aggregation setting on both axes and both datasets. $^\dagger$The majority setting used in Table~\ref{tab:main}. This reduced-scale re-run does not reproduce the extreme clean-accuracy figure of the full-scale majority row in Table~\ref{tab:main} ($5.0\%$ on HotpotQA); we report both rather than reconcile them, and base no claim on that figure.}
\label{tab:robustrag-alpha}
\end{table}

Tuned fairly, RobustRAG is far stronger than Table~\ref{tab:main} suggests. At $\alpha=0.2$ it reaches $6.7\%$/$36.7\%$ (ASR/CleanAcc) on NQ and $3.3\%$/$66.7\%$ on HotpotQA --- competitive with TRIS on NQ and better on both axes on HotpotQA. \textbf{We therefore do not claim ASR dominance over a fairly-tuned RobustRAG.} Two caveats cut in opposite directions: these are $n=30$ against TRIS's $n=100$, and our RobustRAG uses heuristic rather than LLM-based keyword extraction (Section~\ref{sec:setup}).

What does separate the systems at comparable ASR is verification cost: TRIS's L1+L2 default needs one MiniLM pass and one generation call ($\sim\!0.35$\,s/query), against RobustRAG's per-document isolation ($31$--$36$ calls/query) --- a $\sim\!36\times$ reduction. This is regime-specific, not a uniform win: under paraphrased triggers TRIS needs Layer~3, whose cost returns to that range (Section~\ref{sec:adaptive-eval}).

TrustRAG's optional LLM-consistency check --- its closest analogue to Layer~3 --- is likewise disabled in Table~\ref{tab:main}. Enabled, under the adaptive attacks of Section~\ref{sec:adaptive-eval}, the two systems are indistinguishable at $n=30$: every margin is one query wide, with TrustRAG marginally ahead on trigger-paraphrased poisons (Table~\ref{tab:trustrag-adaptive}).

\begin{table}[t]
\centering
\small
\begin{tabular}{lcccc}
\toprule
& \multicolumn{2}{c}{\textbf{TRIS (full)}} & \multicolumn{2}{c}{\textbf{TrustRAG$^\ddagger$}} \\
\cmidrule(lr){2-3}\cmidrule(lr){4-5}
\textbf{Attack} & ASR$\downarrow$ & Clean$\uparrow$ & ASR$\downarrow$ & Clean$\uparrow$ \\
\midrule
Trigger-para. & 27\% & 27\% & 20\% & 33\% \\
Full adaptive & 33\% & 30\% & 30\% & 27\% \\
\bottomrule
\end{tabular}
\caption{\textbf{TRIS vs.\ TrustRAG with its LLM-consistency check enabled} ($n=30$, NQ). $^\ddagger$Unlike Table~\ref{tab:main}, TrustRAG runs with its optional LLM check on. The adaptive poison set differs from Table~\ref{tab:forced-top}, so rows are not comparable across the two. At $n=30$ one query is $3.3$ points: every margin here is one query wide.}
\label{tab:trustrag-adaptive}
\end{table}

\textbf{We therefore withdraw any claim of an adaptive-ASR advantage over a fairly-configured TrustRAG}, and rest the comparison on architecture instead: a dedicated structural layer, an explicit fail-open verdict in Layer~3, and a modular composition whose layers toggle independently at known cost.

A further ablation isolates how much Layer~1's independent embedding geometry contributes. Swapping only that space --- MiniLM versus the retriever's own Contriever, $n=100$, both datasets --- leaves ASR statistically unchanged (NQ $32.0\%$ vs.\ $33.0\%$, $p=0.88$; HotpotQA $44.0\%$ vs.\ $53.0\%$, $p=0.20$), whereas Layer~2 accounts for the large effect ($p<10^{-7}$; Appendix~\ref{sec:appendix-judge}). We report this as a negative result about our own design rationale: the independent geometry is a robustness property --- it drops the assumption that poisons form a separable cluster in the retriever's space --- not the source of the measured gain.

\subsection{Retrieval Dynamics}
\label{sec:retrieval-dynamics}

To understand \emph{how} the Sieve works, Table~\ref{tab:ir} (Appendix~\ref{sec:appendix-ir}) reports retriever-level metrics on NQ. Vanilla poisoning eliminates Recall@5 and drives PoisonedMRR to its maximum of $1.000$ --- a poison at rank~1 for every query. The Sieve fully inverts this: Recall@5 and Clean MRR return to clean-corpus baselines and PoisonedMRR falls to $0.000$. Recall@50 is unchanged throughout, confirming that poisons remain in the candidate pool but no longer reach the visible context: the Sieve reorders rather than discards. TrustRAG achieves the same retriever-side outcome on NQ, consistent with its strong NQ end-to-end numbers; the Sieve's advantage emerges downstream on harder datasets.

\subsection{Layer Ablation}

Table~\ref{tab:ablation} isolates the contribution of each layer on NQ under both black-box and white-box attacks. Three findings stand out.

\paragraph{Layer~2 dominates black-box.} Alone, the structural filter reduces ASR from $67\%$ to $4\%$, capturing the lexical signature of Split-and-Merge prefixes. This is consistent with the design intuition: black-box triggers repeat the query verbatim, producing prefix overlap that is trivially detected.

\paragraph{No single layer suffices for white-box.} Against HotFlip, Layer~2 alone yields little benefit ($\sim\!71\%$), since gradient-optimized triggers evade lexical overlap; only L1+L3 reduces substantially ($27.8\%$). Defeating white-box attacks therefore requires both geometric diversity (Layer~1 disrupts the gradient-targeted geometry) and parametric verification (Layer~3 catches payload contradictions regardless of trigger surface form).

\paragraph{Full system trades black-box for white-box.} Full L1+L2+L3 matches L1+L3's $27.8\%$ white-box ASR but raises black-box ASR ($9.0\%$ vs.\ $3.0\%$ for L1+L2) --- an over-filtering effect, as Layer~3 occasionally discards clean passages whose phrasing differs from the LLM's prior. In practice L1+L2 is preferable for black-box-dominant threat models, with L3 enabled adaptively when white-box adversaries are expected (Section~\ref{sec:discussion}).

\begin{table}[t]
\centering
\small
\begin{tabular}{lccc}
\toprule
\textbf{Config} & \textbf{BB ASR}$\downarrow$ & \textbf{WB ASR}$\downarrow$ & \textbf{Latency} \\
& & & \textbf{(ms/q)} \\
\midrule
No Defense & 67.0\% & $\sim\!74\%^*$ & 0 \\
L1 only & 33.0\% & 57.0\% & 13 \\
L2 only & 4.0\% & $\sim\!71\%^*$ & 8 \\
L3 only & 68.0\% & $\sim\!71\%^*$ & 2{,}948 \\
L1+L2 & \textbf{3.0\%} & 53.0\% & \textbf{12} \\
L1+L3 & 11.0\% & \textbf{27.8\%} & 2{,}286 \\
L2+L3 & 3.0\% & 68.0\% & 154 \\
Full (L1+L2+L3) & 9.0\% & 27.8\% & 15{,}202 \\
\bottomrule
\end{tabular}
\caption{\textbf{Layer ablation on NQ ($k=50$).} BB = black-box LM-targeted; WB = white-box HotFlip. $^*$WB rows marked $\sim$ are computed over valid responses only (see Section~\ref{sec:setup}). L2 alone dominates black-box; only combinations including L1 and L3 meaningfully reduce white-box ASR. Full system trades black-box ASR for white-box robustness due to over-filtering by Layer~3.}
\label{tab:ablation}
\end{table}

\paragraph{HotpotQA supplementary ablation.} On multi-hop HotpotQA, L2+L3 attains the lowest black-box ASR ($11.0\%$), just below L1+L2 ($12.0\%$), while the full system rises to $31.0\%$ --- again over-filtering when L3 is applied across multi-hop evidence. We treat this as a tunable deployment choice, not a fundamental limitation.

\paragraph{Layer orthogonality.} Three observations from Table~\ref{tab:ablation} support orthogonal rather than redundant coverage. L1 alone leaves $33\%$ black-box ASR and L2 alone $4\%$, yet L1+L2 reaches $3\%$ --- L1 catches paraphrasing poisons L2 misses. In the white-box regime L1 alone leaves $57.0\%$ and L3 alone leaves $\sim\!71\%$, yet L1+L3 reaches $27.8\%$, so their catches barely overlap. And Full ($9\%$) is \emph{worse} than L1+L2 ($3\%$): the layers are not strictly additive, because L3's false-positive rate on clean documents matters. They therefore cover distinct failure modes while interacting in false-positive profiles --- supporting the recommended configuration over a naive ``run all three.''

\subsection{Adaptive Adversary Evaluation} \label{sec:adaptive-eval}

Section~\ref{sec:adaptive-threat} defines four levels of attacker adaptivity. This section empirically evaluates Level 1 (architecture-aware); Levels 2 (judge-aware) and 3 (full white-box) remain conceptual only, and implementing them is planned future work.

\subsubsection{Paraphrasing evades Layer 2 without sacrificing retrieval rank}

Letting the live Contriever retriever rank poisons ($k=50$, $n=100$), paraphrasing the trigger costs the attacker almost nothing at retrieval: $4.97$ of $5$ poisons still reach the top-$50$ on NQ (versus $5.00$ verbatim) and $5.00$ of $5$ on HotpotQA, and undefended ASR does not consistently drop (NQ $55.0\%\!\to\!45.0\%$; HotpotQA $70.0\%\!\to\!77.0\%$). Paraphrasing is thus a free evasion of Layer 2's lexical check, shifting the burden to Layers 1 and 3. Layer 1 alone recovers much of the loss ($45.0\%\!\to\!32.0\%$ on NQ, $77.0\%\!\to\!44.0\%$ on HotpotQA), matching the Layer-3-off rows of Table~\ref{tab:adaptive-live}. That $32.0\%$ is within noise of Layer 1's $33.0\%$ verbatim-trigger ASR in Table~\ref{tab:ablation}, so its geometric check is largely insensitive to surface paraphrasing, as its design predicts. A verbatim-trigger sieve reproduces Table~\ref{tab:main}'s defended NQ ASR exactly ($3.0\%$), though the undefended baseline in this live-retrieval setup (NQ $55.0\%$, HotpotQA $70.0\%$) differs somewhat from Table~\ref{tab:main}'s static baseline (NQ $67.0\%$, HotpotQA $87.0\%$), likely reflecting live vs.\ cached retrieval.

\subsubsection{Enabling Layer 3 recovers the loss on all three axes at once}

Holding Layer 1 and Layer 2 fixed across paired runs (Layer 1 removes $17.05$ (NQ) / $14.47$ (HotpotQA) documents per query on average; Layer 2 removes $0.01$ / $0.15$), we toggle Layer 3 alone against paraphrased-trigger poisons (Table~\ref{tab:adaptive-live}).

\begin{table*}[t]
\centering
\small
\begin{tabular}{lcccccc}
\toprule
& \multicolumn{3}{c}{\textbf{NQ}} & \multicolumn{3}{c}{\textbf{HotpotQA}} \\
\cmidrule(lr){2-4} \cmidrule(lr){5-7}
\textbf{Layer 3} & ASR$\downarrow$ & CleanAcc$\uparrow$ & Surv./5$\downarrow$ & ASR$\downarrow$ & CleanAcc$\uparrow$ & Surv./5$\downarrow$ \\
\midrule
Off & 32.0\% & 43.0\% & 2.51 & 44.0\% & 34.0\% & 2.55 \\
On  & \textbf{15.0\%} & \textbf{61.0\%} & \textbf{0.79} & \textbf{31.0\%} & \textbf{45.0\%} & \textbf{1.22} \\
\bottomrule
\end{tabular}
\caption{\textbf{Layer 3 under live retrieval.} Paraphrased-trigger poisons, Contriever, $k=50$, $n=100$; Layers 1--2 held fixed across the paired runs, so differences are attributable to Layer 3 alone. ``Surv./5'' is the mean number of the 5 injected poisons still in context after filtering. Enabling Layer 3 improves all three columns simultaneously on both datasets.}
\label{tab:adaptive-live}
\end{table*}

This is not an ASR-for-utility trade: on NQ, ASR roughly halves while clean accuracy rises $18$ points. The honest cost is latency --- Layer 3's per-document verdicts add $\sim\!16$--$19$ s/query under live retrieval against a $\sim\!0.35$ s L1+L2 baseline --- though adaptive invocation, firing only when Layers 1 and 2 disagree, keeps this occasional ($21\%$ of HotpotQA queries). We therefore position Layer 3 as an optional high-assurance layer rather than part of the black-box-dominant default (Section~\ref{sec:defense}).

\subsubsection{Forced-top worst case}

As a harsher upper bound we pin poisons to the top of context, removing retrieval competition, and ablate Level 1 adaptivity into \emph{payload-diversified}, \emph{trigger-paraphrased}, and \emph{full adaptive} variants against the static baseline ($n=30$, NQ; Table~\ref{tab:forced-top}, Appendix~\ref{sec:appendix-forced-top}). Layer 2 neutralizes the static and payload-diversified attacks ($10.0\%$) at near-zero cost, since neither alters the prefix it inspects; paraphrasing the trigger evades it ($57.0\%$ under L1+L2) but forces a claim contradicting the model's prior, which Layer 3 catches ($40.0\%$ and $33.0\%$).

\subsection{Injection Ratio and Hyperparameter Sweeps}
\label{sec:injection-sweep}
\label{sec:sensitivity}

Varying poisons per query from $1$ to $5$ on NQ (Table~\ref{tab:inject}, Appendix~\ref{sec:appendix-inject}), baseline ASR climbs from $0\%$ to $66\%$ while the Sieve holds below $10\%$; extending to $6$--$20$ poisons ($n=25$; Table~\ref{tab:inject-ext}) leaves TRIS ASR flat at $4.0\%$. Because Layers 1 and 2 filter independently rather than by majority vote, denser poisoning cannot tip the defense the way it could tip a voting aggregator. Full-scale replication remains future work (Section~\ref{sec:limits}).

Sweeping retrieval depth $k \in \{5,10,20,50\}$ against cluster count $C \in \{2,3,5\}$ (Table~\ref{tab:heatmap}, Appendix~\ref{sec:appendix-heatmap}), the defense is unstable at $k=5$ (ASR $85$--$93\%$, clean accuracy $9$--$17\%$) because K-Means partitions on too few candidates are unreliable. From $k=10$ onward ASR stabilizes at $3$--$4\%$ and clean accuracy rises monotonically to $78\%$ at $k=50$; $C=3$ is our default, and $C=5$ performs comparably on this sweep --- we fix $C=3$ as the more conservative setting.

\subsection{Latency}
\label{sec:cases}

Layer~2 adds $\sim\!8$\,ms/query and Layer~1 $\sim\!13$\,ms; Layer~3 dominates at $\sim\!2.9$\,s/query, making always-on ($\sim\!15$\,s) acceptable for high-risk queries but not high-throughput search --- under live retrieval its added cost is similar ($\sim\!16$--$19$\,s/query; Section~\ref{sec:adaptive-eval}). Appendix~\ref{sec:appendix-case-studies} traces one successful defense and one failure, showing where the residual HotpotQA ASR originates.

%-------------------------------------------------------------------------------
\section{Discussion}
\label{sec:discussion}
%-------------------------------------------------------------------------------

\paragraph{Adaptive attacks on the judge.} Section~\ref{sec:adaptive-eval} closes the Level 1 gap empirically; judge-awareness and white-box access remain conceptual. Against a judge-aware attacker (Level 2) Layer~1 becomes a single point of failure; ensembling diverse judges and rotating $\mathcal{J}$ are the natural countermeasures, and since Layer~1's embedding-space independence alone does not measurably differentiate attack outcomes (Appendix~\ref{sec:appendix-judge}), ensembling multiple judges is the more promising direction. A Level 3 adversary also knows Layer~2's thresholds and could search for payloads at the decision boundary of $y_{\mathrm{int}}$; randomizing that threshold per query and ensembling verifiers are analogous responses, both future work.

\paragraph{Zero-Trust Retrieval.} TRIS treats retrieved evidence as untrusted input requiring active validation, analogous to a network firewall --- which is also why it sits in middleware rather than retraining the retriever: adversarial training is expensive, must be redone as attacks emerge, and is incompatible with third-party embedding APIs, whereas middleware is corpus-, model-, and vendor-agnostic. Complementary modules could address prompt injection \citep{greshake2023notwhat}, privacy leakage, and content moderation under the same principle, keeping the utility--robustness trade-off explicit through adaptive invocation.

%-------------------------------------------------------------------------------
\section{Conclusion}
%-------------------------------------------------------------------------------

We presented TRIS, a three-layer middleware defense against retrieval-stage poisoning, each layer attacking a distinct constraint a poisoned document must satisfy. It reduces black-box ASR by an order of magnitude on NQ and MS-MARCO, mitigates white-box HotFlip from $\sim\!74\%$ to $27.8\%$ with Layer~3 enabled, and drives poisoned MRR to zero, while recovering $41$--$45$ points of clean accuracy over the attacked baseline. Against fairly-tuned TrustRAG and RobustRAG it is competitive rather than dominant (Section~\ref{sec:fair-baselines}); what distinguishes it is a strong worst-case profile, far lower verification cost in the common attack regime, and graceful degradation as attackers adapt.

%-------------------------------------------------------------------------------
\section*{Limitations}
\label{sec:limits}
%-------------------------------------------------------------------------------

\paragraph{No empirical evaluation of judge-aware or white-box adversaries.} A remaining limitation of this work is that our headline numbers (Section \ref{sec:results}) are against static PoisonedRAG attacks (Level 0 in Section~\ref{sec:adaptive-threat}). Section~\ref{sec:adaptive-eval} now provides an empirical evaluation of Level 1 (architecture-aware) adversaries --- paraphrased triggers and diversified payloads, under live retrieval and a forced-top worst case --- showing that Layer 3 recovers most of the robustness that paraphrasing costs Layer 2. Levels 2 (judge-aware) and 3 (full white-box) remain conceptual only (Section~\ref{sec:adaptive-threat}); implementing them is planned future work. An adversary who optimizes triggers to land in the majority cluster of the judge model, avoid prefix overlap with the query, and produce payloads compatible with the LLM's parametric prior would represent the worst case for TRIS. Whether such an attacker can simultaneously satisfy all three constraints --- and how much trigger-and-payload search space they would need to explore --- remains an open empirical question for Levels 2--3 that follow-up work should address with adaptive-attack benchmarks of the kind suggested by \citet{ma2025benchmarking}.

% The most significant limitation of this work is that all reported numbers are against static PoisonedRAG attacks (Level 0 in Section~\ref{sec:adaptive-threat}). We provide a conceptual analysis of architecture-aware (Level 1), judge-aware (Level 2), and full white-box (Level 3) adversaries, but we do not implement and evaluate them. An adversary who optimizes triggers to land in the majority cluster of the judge model, avoid prefix overlap with the query, and produce payloads compatible with the LLM's parametric prior would represent the worst case for TRIS. Whether such an attacker can simultaneously satisfy all three constraints --- and how much trigger-and-payload search space they would need to explore --- is an open empirical question that follow-up work should address with adaptive-attack benchmarks of the kind suggested by \citet{ma2025benchmarking}.

\paragraph{Generator dependency and Layer 3 coupling.} Layer~3's contradiction detection is only as reliable as the verifier's own parametric knowledge of $y_{\mathrm{true}}$ --- a knowledge-dependency that is a structural property of the design, not an artifact of any one model. Our headline results use GPT-3.5-turbo as both the generator and (in Layer~3 always-on mode) the verifier. This couples the defense's measured efficacy to the parametric knowledge of a specific commercial model. We tested this failure mode directly on a slice of $10$ questions about clearly post-cutoff events the generator does not know: Layer~3 abstained on all $10$ (fail-open), no clean document was wrongly dropped ($0$ of $50$), and Layers~1--2 still removed every injected poison ($30$ of $30$). Layer~3 therefore degrades to a no-op rather than to a source of false positives when its parametric prior is absent, and the structural layers carry the defense unaided --- consistent with the L1+L2 row of Table~\ref{tab:ablation}, which reaches $3.0\%$ black-box ASR with Layer~3 disabled entirely.

\paragraph{Scale and corpus diversity.} We evaluate on $100$-query subsets of NQ, HotpotQA, and MS-MARCO. Recall@5 on the clean baseline is $0.180$, reflecting the subset rather than full-corpus retrieval. Scaling to full corpora (BEIR, full MS-MARCO, enterprise document stores) is essential for characterizing false-positive rates in production and is the most pressing follow-up.

\paragraph{Statistical reporting.} Tables report point estimates over $10$ iterations $\times$ $10$ queries. Differences of $\leq\!3$ percentage points (e.g., Sieve vs.\ TrustRAG on HotpotQA) may be within iteration-to-iteration noise; we frame those comparisons as ``comparable'' rather than wins.

\paragraph{Cost of Layer 3.} The LLM consistency layer adds $\sim\!2.9$\,s/query and is the dominant latency cost. Distilling LLM-as-judge behavior into a smaller specialized verifier is the obvious next step; the adaptive-invocation mode is a stopgap.

\paragraph{Injection-ratio coverage.} Our injection sweep covers $1$--$5$ adversarial documents per query at full scale ($n=100$), where the baseline ASR climbs from $0\%$ to $66\%$ and the Sieve holds below $10\%$. The Sieve's ASR is non-monotonic in this range ($8\% \to 10\% \to 6\% \to 8\%$), which we attribute to iteration noise at small sample sizes. We extend this range to $6$--$20$ adversarial documents per query on a reduced-scale NQ subsample ($n=25$; Table~\ref{tab:inject-ext}): TRIS ASR stays flat at $4.0\%$ and CleanAcc remains near $46\%$, supporting the``robustness to denser poisoning'' hypothesis. Full-scale replication of this extended range, and coverage of HotpotQA and MS-MARCO, remains future work.

% Our injection sweep covers $1$--$5$ adversarial documents per query, where the baseline ASR climbs from $0\%$ to $66\%$ and the Sieve holds below $10\%$. The Sieve's ASR is non-monotonic in this range ($8\% \to 10\% \to 6\% \to 8\%$), which we attribute to iteration noise at small sample sizes; this also signals that statements about ``robustness to denser poisoning'' should be tested at $6$--$20$ adversarial documents per query, a range we were unable to complete due to API quotas. We do not expect qualitative changes but the hypothesis is untested.

\paragraph{Reduced-scale baseline comparisons.} The fair-baseline re-evaluations that our comparative claims now rest on --- the RobustRAG threshold sweep (Table~\ref{tab:robustrag-alpha}) and the TrustRAG comparison with its LLM check enabled (Table~\ref{tab:trustrag-adaptive}) --- were run at $n=30$, against $n=100$ for Table~\ref{tab:main}, and as single runs rather than $10$-iteration aggregates. They are adequate to locate each baseline's operating-point frontier, which is what they are used for, but not to resolve differences of a few points; the $n=30$ RobustRAG majority row also does not reproduce the full-scale majority row in Table~\ref{tab:main}, and we have not been able to reconcile the two. Re-running both baselines at full scale is the first item of follow-up work. We note the direction of the resulting bias honestly: these are the comparisons in which the baselines look \emph{strongest} relative to TRIS, so under-powering them is not a limitation that flatters us.

\paragraph{No formal guarantees.} TRIS is heuristic. It does not provide certified robustness in the sense of \citet{xiang2024robustrag} and we make no formal claim about adversarial bounds. Combining TRIS's empirical strength with certified isolate-and-aggregate decoding is a promising direction --- e.g., applying TRIS as a pre-filter and RobustRAG as the certified aggregator.

\paragraph{Partial-iteration results.} The white-box L1+L3 and full-system numbers in Table~\ref{tab:ablation} are averaged over $9$ of $10$ iterations, because one backup job did not complete (unrelated to the transient API failures in Section~\ref{sec:setup}) and we did not have the compute budget to re-run it. Judging by the iteration-to-iteration spread elsewhere in our runs, we expect these two numbers to move by at most $1$--$2$ percentage points; we report them as partial rather than complete, and no claim in this paper turns on a margin that small.

%-------------------------------------------------------------------------------
\section*{Ethical Considerations}
%-------------------------------------------------------------------------------

This work studies a defense against an existing, publicly documented attack class (PoisonedRAG and its follow-ups). We do not introduce new attack capabilities. The defense is designed to be deployable as a middleware module by RAG operators without requiring retriever or generator retraining, lowering the barrier to robust deployment. Our experiments use publicly available datasets (NQ, HotpotQA, MS-MARCO) and do not involve human subjects or personally identifiable information. We acknowledge a dual-use consideration: detailed descriptions of attacker design (Section~3) inform defenders but could in principle aid attackers; however, all attack details follow \citet{zou2025poisonedrag} and are already public. We believe the net effect of clearer, comparable defense evaluations is positive for the security of deployed RAG systems.

%-------------------------------------------------------------------------------
\section*{Acknowledgments}
%-------------------------------------------------------------------------------

This work is based upon the work supported by the National Center for Transportation Cybersecurity and Resiliency (TraCR) (a U.S. Department of Transportation National University Transportation Center) headquartered at Clemson University, Clemson, South Carolina, USA. It was also supported in part by NIST grant number 60NANB24D143. Any opinions, findings, conclusions, and recommendations expressed in this material are those of the author(s) and do not necessarily reflect the views of NIST, TraCR, and the U.S. Government assumes no liability for the contents or use thereof.

This work used the Delta system at the National Center for Supercomputing Applications [award OAC 2005572] through allocation CIS251331 from the Advanced Cyberinfrastructure Coordination Ecosystem: Services \& Support (ACCESS) program, which is supported by National Science Foundation grants \#2138259, \#2138286, \#2138307, \#2137603, and \#2138296. The authors also acknowledge High Performance Computing at The University of Texas at Dallas (HPC@UTD) for providing computing resources.

Disclaimer: This paper identifies certain equipment, instruments, software, or materials to adequately describe the experimental procedure. Such identification is not intended to imply recommendation or endorsement of any product or service by NIST, nor is it intended to imply that the materials or equipment identified are necessarily the best available for the purpose.

\bibliography{references}

\appendix

\section{Threat Model Illustration}
\label{sec:appendix-fig1}

\begin{figure*}[b]
    \centering
    \includegraphics[width=0.95\textwidth]{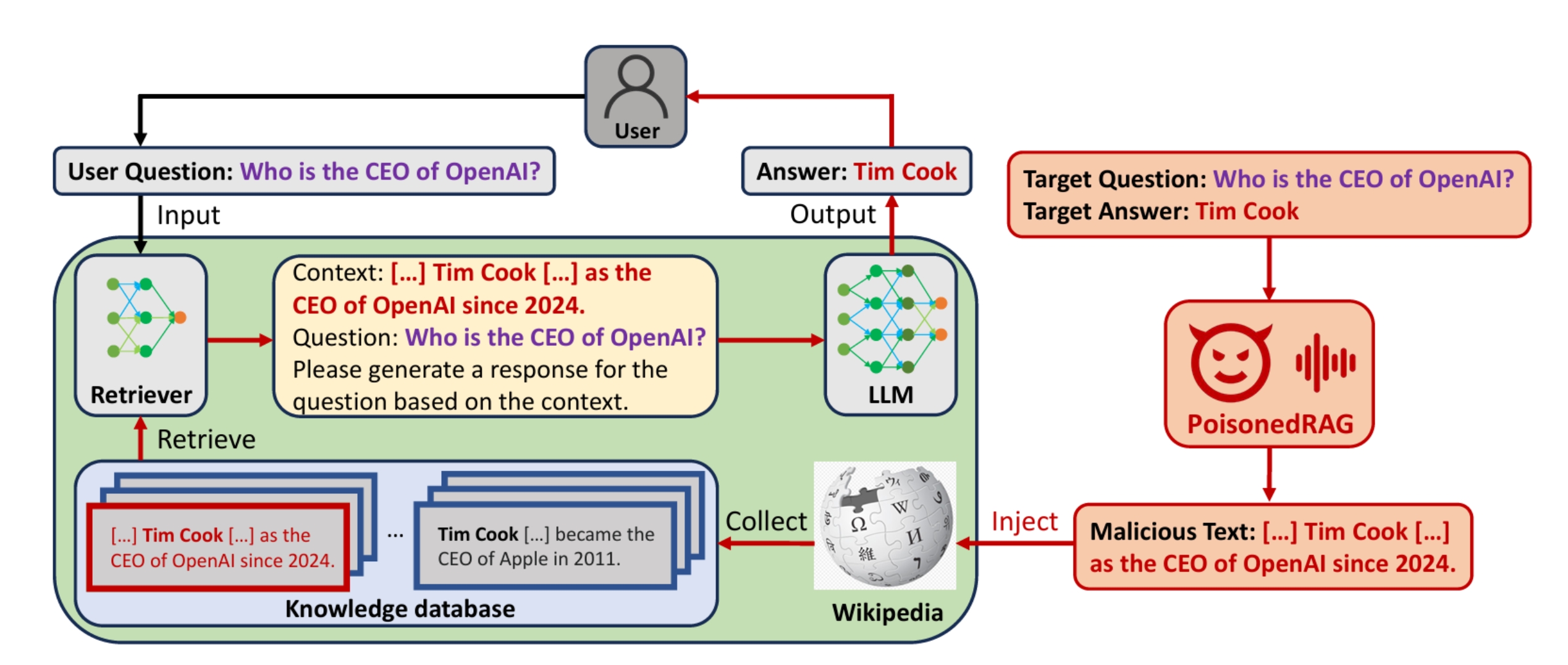}
    \caption{\textbf{The PoisonedRAG threat model.} The attacker injects a small number of adversarially crafted documents into a large retrieval corpus. Dense retrievers \citep{karpukhin2020dense,izacard2021contriever} rank these poisoned documents highly for the target query due to trigger optimization. The LLM then conditions on malicious context and outputs the attacker-specified answer \citep{zou2025poisonedrag}.}
    \label{fig:poisonedrag-threat-model}
\end{figure*}

Section~\ref{sec:threat-model} describes the PoisonedRAG threat model in prose: an adversary injects a small number of adversarially crafted documents into the retrieval corpus, a dense retriever ranks them highly for the target query because of trigger optimization, and the generator conditions on this poisoned context and outputs the attacker's chosen answer instead of the true one. Figure~\ref{fig:poisonedrag-threat-model} gives the visual walkthrough of that pipeline for a single example query, from injection through retrieval to the corrupted final answer; it is reproduced here rather than in Section~\ref{sec:threat-model} to keep the main body within the page limit.

\section{Extended Results}
\label{sec:appendix-results}

This appendix collects the full tables behind results that Section~\ref{sec:results} reports and interprets in prose: retrieval-level diagnostics, the adaptive-adversary breakdown, the extended injection-ratio sweep, and the retrieval-depth/cluster-count sensitivity sweep. None of the numbers below are new --- each table is cited from the point in the main text where its findings are first discussed, and is reproduced here, rather than inline, to keep the main body within the page limit. Each subsection also adds a short pointer back to the relevant main-text discussion.

\subsection{Retrieval Dynamics}
\label{sec:appendix-ir}

Table~\ref{tab:ir} is the retriever-level counterpart to Table~\ref{tab:main}: instead of end-to-end accuracy and attack success, it reports Recall@5, Recall@50, and MRR directly on NQ, isolating how the Sieve changes the ranking of retrieved documents rather than only the generator's final answer. Discussed in Section~\ref{sec:retrieval-dynamics}.

\begin{table}[!h]
\centering
\setlength{\tabcolsep}{4pt}
% \small
\resizebox{\columnwidth}{!}{%
\begin{tabular}{lcccc}
\toprule
\textbf{System State} & \textbf{R@5}$\uparrow$ & \textbf{R@50}$\uparrow$ & \textbf{CleanMRR}$\uparrow$ & \textbf{PoisMRR}$\downarrow$ \\
\midrule
Clean Corpus & 0.180 & 0.680 & 0.131 & --- \\
Poisoned (No Defense) & 0.000 & 0.680 & 0.048 & 1.000 \\
{} + TrustRAG & 0.180 & 0.680 & 0.131 & 0.000 \\
{} + Tri-Layer Sieve & \textbf{0.180} & \textbf{0.680} & \textbf{0.131} & \textbf{0.000} \\
\bottomrule
\end{tabular}%
}
\caption{\textbf{Retrieval dynamics on NQ, top-50.} The Sieve fully inverts the MRR corruption induced by poisoning: PoisonedMRR drops from $1.000$ to $0.000$ while Recall@5 and CleanMRR return to clean-corpus baseline. Recall@50 is unaffected, confirming that the defense reorders candidates rather than discarding the candidate pool.}
\label{tab:ir}
\end{table}
% \FloatBarrier

\subsection{Forced-Top Worst Case}
\label{sec:appendix-forced-top}

Table~\ref{tab:forced-top} reports the forced-top ablation discussed in Section~\ref{sec:adaptive-eval}, in which poisons are pinned to the top of context to remove retrieval competition and isolate the defense's behavior under the harshest plausible placement, separating the payload-diversified and trigger-paraphrased components of Level~1 adaptivity before combining them.
\FloatBarrier

\begin{table}[!h]
\centering
\small
\begin{tabular}{lccc}
\toprule
\textbf{Attack} & \textbf{Undef.} & \textbf{L1+L2} & \textbf{Full} \\
\midrule
Black-box (static) & 70.0\% & 10.0\% & 10.0\% \\
Payload-div.        & 67.0\% & 10.0\% & 10.0\% \\
Trigger-para.        & 67.0\% & 57.0\% & 40.0\% \\
Full adaptive        & 70.0\% & 50.0\% & 33.0\% \\
\bottomrule
\end{tabular}
\caption{\textbf{Forced-top worst case.} Poisons pinned to the top of context ($n=30$, NQ, ASR$\downarrow$) --- an upper bound harsher than the live-retrieval Table~\ref{tab:main} (TRIS NQ ASR $3.0\%$). Layer 2 alone stops static and payload-diversified attacks; Layer 3 is needed once the trigger is paraphrased.}
\label{tab:forced-top}
\end{table}
\FloatBarrier

\subsection{Does Layer 1's Independent Geometry Matter?}
\label{sec:appendix-judge}

Section~\ref{sec:defense} frames Layer~1 as majority-keep in an architecturally independent geometry. This appendix reports the ablation that tests whether that independence is itself load-bearing, summarized in Section~\ref{sec:fair-baselines}. We swap only Layer~1's embedding space --- Sentence-BERT MiniLM (architecturally independent of the retriever) versus Contriever (the retriever's own space) --- holding every other component fixed at $k=50$, $n=100$ on NQ and HotpotQA.

Under paraphrased-trigger attacks the two spaces are statistically indistinguishable: NQ ASR $32.0\%$ (MiniLM) versus $33.0\%$ (Contriever), $p=0.88$; HotpotQA $44.0\%$ versus $53.0\%$, $p=0.20$; clean-accuracy differences are likewise not significant. The component that does differentiate is Layer~2: on the verbatim-trigger attack, adding it drives ASR to $3.0\%$ (NQ) and $8.0\%$ (HotpotQA) against $33.0\%$ and $44.0\%$ for clustering alone ($p<10^{-7}$).

We report this as a negative result about our own design rationale rather than omitting it. Layer~1's independent geometry remains a robustness property --- it removes the requirement that poisons form a tight, separable cluster in the retriever's own space, an assumption that a judge-aware adversary would target directly --- but it is not where the measured gain comes from, and we scope our novelty claim to Layer~2 and the orthogonal composition accordingly. Judge ensembling and rotation (Section~\ref{sec:discussion}) remain plausible routes to making that independence pay off empirically; we have not tested them.

\raggedbottom 
\subsection{Injection Ratio Sweeps}
\label{sec:appendix-inject}

Tables~\ref{tab:inject} and~\ref{tab:inject-ext} report the full injection-ratio sweep discussed in Section~\ref{sec:injection-sweep}. Table~\ref{tab:inject} covers $1$--$5$ adversarial documents per query at full scale ($n=100$); Table~\ref{tab:inject-ext} extends this to $6$--$20$ documents per query on a reduced-scale sample ($n=25$) to check whether the Sieve's robustness holds at higher poisoning density.

\begin{table}[!h]
\centering
\small
\begin{tabular}{ccc}
\toprule
\textbf{Adv. docs/query} & \textbf{Baseline ASR} & \textbf{Sieve ASR} \\
\midrule
1 & 0.0\% & 0.0\% \\
2 & 47.0\% & 8.0\% \\
3 & 54.0\% & 10.0\% \\
4 & 64.0\% & 6.0\% \\
5 & 66.0\% & 8.0\% \\
\bottomrule
\end{tabular}
\caption{\textbf{Injection-ratio sweep on NQ ($k=50$).} The Sieve holds ASR below $10\%$ as the baseline climbs from $0\%$ to $66\%$.}
\label{tab:inject}
\end{table}

\begin{table}[H]
\centering
\small
\begin{tabular}{lcccc}
\toprule
\textbf{Poisons/query} & \textbf{6} & \textbf{10} & \textbf{15} & \textbf{20} \\
\midrule
Baseline ASR$\downarrow$ & 52.0\% & 48.0\% & 52.0\% & 56.0\% \\
Sieve ASR$\downarrow$    & 4.0\%  & 4.0\%  & 4.0\%  & 4.0\%  \\
Sieve CleanAcc$\uparrow$ & 48.0\% & 44.0\% & 48.0\% & 48.0\% \\
\bottomrule
\end{tabular}
\caption{\textbf{Injection sweep, extended range.} $6$--$20$ adversarial documents/query, NQ, $n=25$ (cf.\ Table~\ref{tab:inject} for $1$--$5$ at full scale, $n=100$). ASR stays flat regardless of poison density.}
\label{tab:inject-ext}
\end{table}
\FloatBarrier

\subsection{Sensitivity to Retrieval Depth and Cluster Count}
\label{sec:appendix-heatmap}

Table~\ref{tab:heatmap} reports the full retrieval-depth ($k$) by cluster-count ($C$) sweep summarized in Section~\ref{sec:sensitivity}.

\begin{table}[H]
\centering
\small
\begin{tabular}{lccc}
\toprule
\textbf{$k$ $\backslash$ $C$} & \textbf{$C{=}2$} & \textbf{$C{=}3$} & \textbf{$C{=}5$} \\
\midrule
$k{=}5$ & --- & 93\% / 9\% & 85\% / 17\% \\
$k{=}10$ & 4\% / 69\% & 4\% / 61\% & 3\% / 69\% \\
$k{=}20$ & 4\% / 71\% & 3\% / 71\% & 3\% / 72\% \\
$k{=}50$ & 4\% / 75\% & \textbf{3\% / 77\%} & 3\% / 78\% \\
\bottomrule
\end{tabular}
\caption{\textbf{Sensitivity to $k$ and $C$ on NQ.} Each cell is ASR\,/\,CleanAcc. The $k{=}5,C{=}2$ cell failed due to an API error and is omitted. $k\geq 10$ yields stable ASR--utility; $C=3$ is our default, with $C=5$ performing comparably on this sweep.}
\label{tab:heatmap}
\end{table}
\FloatBarrier

\section{Qualitative Case Studies}
\label{sec:appendix-case-studies}

Section~\ref{sec:cases} reports aggregate latency, and Table~\ref{tab:main} reports aggregate attack-success and clean-accuracy rates. Aggregate rates do not show \emph{why} the defense succeeds on most queries or \emph{how} it fails on the rest, so this appendix walks through one query of each kind in detail: one where the Sieve correctly removes the poison, and one where a poison survives all three layers and the attack succeeds.

\paragraph{Success (black-box).} For an NQ query about a technology executive, five poisons each begin with the verbatim query followed by an authoritative paragraph naming a plausible alternative entity. The poisons rank in the top-$5$ under Contriever; in judge space they form a tight off-cluster (Layer~1 flags four of five) and Layer~2 independently flags all five via $>\!0.9$ Jaccard overlap. The generator emits the correct answer.

\paragraph{Failure (mimicry).} For a HotpotQA two-hop query, a poison mimics the genuine Wikipedia passage's style, differing only in the answer entity. It does not repeat the query (evading Layer~2) and clusters with benign passages in judge space (evading Layer~1); Layer~3's signal is weak because the alternative entity is itself plausible. This pattern accounts for most of the residual $14\%$ HotpotQA ASR reported in Table~\ref{tab:main}: multi-hop amplifies mimicry failures because a poison need only corrupt one hop.

Residual risk is concentrated in payload-level mimicry rather than trigger evasion --- the failure profile a stronger Layer~3 would address.

\end{document}